# Effectiveness and Limitations of Statistical Spam Filters

**M. Tariq Banday,** *Lifetime Member, CSI*
*P.G. Department of Electronics and Instrumentation Technology*
*University of Kashmir, Srinagar, India*

**Tariq R. Jan**
*P. G. Department of Statistics*
*University of Kashmir, Srinagar, India*

**Abstract**

Spam is not only clogging the Internet traffic by consuming a hefty amount of network bandwidth but it is also a source for e-mail born viruses, spyware, adware and Trojan Horses. It is also used to carry out denial of service, directory harvesting and phishing attacks that directly cause financial losses. Further, the contents of spam are often offensive and contain adult oriented and fraudulent materials which are objectionable to recipients. Several anti-spam procedures are currently employed to distinguish spam from legitimate e-mails; however spammers and phishers employ dynamic spam structures to obfuscate e-mail content to circumvent these procedures. Apart from other technological procedures various adaptive learning filters have been developed that have an ability to allow an algorithm to constantly learn what sort of e-mail's or e-mail content a recipient would typically process and what to see in normal course of its business. These filters are based on complex statistical techniques that classify future e-mails based on the word content of accepted e-mails. The statistical techniques employed in these filters separate an incoming e-mail into tokens and assign a probability value to each token. The probability of each token are collectively used to calculate the overall spam probability and accordingly the incoming e-mail is scored as spam, probably spam or legitimate e-mail.

In this paper we discuss the techniques involved in the design of the famous statistical spam filters that include Naïve Bayes, Term Frequency-Inverse Document Frequency, K-Nearest Neighbor, Support Vector Machine, and Bayes Additive Regression Tree. We compare these techniques with each other in terms of accuracy, recall, precision, etc. Further, we discuss the effectiveness and limitations of statistical filters in filtering out various types of spam from legitimate e-mails.

1. Introduction

Despite the development and deployment of various state-of-art anti-spam procedures, spam constitutes more than 75% of total e-mail messages [1]. Spam causes several problems [2] either directly or indirectly to the e-mail system that include: i) Network conjunction, ii) misuse of storage space and computational resources, iii) loss of work productivity and annoyance to users, iv) legal issues as a result of pornographic advertisements and other objectionable material, v) financial losses through phishing and other related attacks, vi) spread of viruses, worms and Trojan Horses, and vii) Denial of Services and Directory Harvesting attacks. Spam is injected at various places into the e-mail system by spammers for their illicit financial gains using a variety of techniques [3] and tools that include spoofing, botnets, open proxies, mail relays, bulk mail tools called mailers, etc. Spammers are proactive and use dynamic spam structures that constantly keep on changing the spam structure to circumvent anti-spam procedures. The increase in the use of anti-spam procedures has led spammers to push more spam into the system in order to reach more and more users and guarantee themselves a huge profit. Several anti-spam procedures have been proposed that try to tackle the problem of spam at various levels in the system [4]. These procedures propose the use of diverse technological, legal, social and economical solutions. According to research work carried out in [5], filtering approaches are currently applied at a wider scale owing to the reason that they do not require any infrastructural change in the existing e-mail system. There have been several approaches in the design of the spam filters namely learning-based filtering, rule-based filtering, non-content filtering, collaborative filtering and hybrid filtering. Among these, learning-based approaches which are based on some statistical technique have been most effective in spam filtering.

This paper is organized as follows: section 2 introduces various statistical techniques used in filtering spam from the legitimate e-mail. In section 3 and 4 limitations of spam filters and the metrics used for evaluating a spam filter are presented. In Section 5, which is the main contribution of this paper, we present the experimental evaluation of four spam filters each of which is based on Naïve Bayesian, K-Nearest Neighbor, Support Vector Machine and CBART classification techniques. We discuss the comparative performance of these spam filters. Finally, we conclude in section 6 and present the future outlook.

2. Statistical Spam Filter Techniques

An e-mail message consists of two parts, namely header and body. Header is a structured set of fields, each having a name and specific meaning. It includes fields namely From, To, Subject, CC, BCC, etc. Message body is usually text, possibly with HTML markup and generally referred to as content of the message. Message analysis and filtering involves selection of features from header and/or body or from message as a whole relevant for analysis. A filter may check the presence of certain words or may consider the arrival of a dozen of substantially identical messages in a certain slot of time. In addition to this, a learning-based filter analyzes a collection of labeled training data which are pre-collected messages with reliable judgment.

A spam filter in general is an application that implements a function to classify an incoming e-mail message as spam or legitimate mail using a particular classification method. Such a system implements the following function:

$$f(m, \theta) = \begin{cases} C_{spam}, & \text{if classified as SPAM} \\ C_{leg}, & \text{otherwise} \end{cases}$$

where, $m$ is the message to be classified, $\theta$ is a vector of parameters, and $C_{spam}$ and $C_{leg}$ are respectively spam and legitimate messages.

Most of the Spam filters including the statistical spam filters use machine learning classification techniques wherein the vector of parameter $\theta$ is the result of training the classifier on a pre-collected dataset which may be rebuilding itself with every new message. $\theta$ for such filters can be defined as:

$$\theta = \Theta(X),$$

Learning-based spam filters treat the input data as an unstructured set of tokens, filtering can be applied either to the whole message or to any part of it. For this group of filters with two classes of messages: spam and legitimate mail, there exists a set of labeled training messages, each message being a vector of $d$ binary features and each label being $C_{spam}$ or $C_{leg}$ depending on the class of the message. The training data set M, once pre-processed in this way, can be described as:

$$X = \{(\bar{x}_1, y_1), (\bar{x}_2, y_1), \dots (\bar{x}_n, y_n)\},$$

$$\bar{x}_i \in \mathbb{Z}_2^d, \qquad y_i \in \{C_{spam}, C_{leg}),$$

where, $d$ is the number of features used. $\bar{x}_i \in \mathbb{Z}_2^d$ is a new sample the classifier should provide a decision $y \in \{C_{spam}, C_{leg}\}$. $y_1, y_2, \dots y_n$ and labels and $\Theta$ being the training function.

*A. Naïve Bayesian Classifier*

Naïve Bayesian classifier [6, 7] is based on Baye's theorem and the theorem of probability. This classifier can be considered as improved learning based variant of keyword filtering when applied to text classification. It uses naive independence assumption that all the features are statistically independent. The basic decision rule used in this classifier can be defined as follows:

$$f(\bar{x}) = \underset{y=\{C_{spam}, C_{leg}\}}{argmax} \left( \hat{p}(y) \prod_{j:x^j=1} \hat{p}(x^j = 1|y) \right),$$

were $x^j$ is the $j^{th}$ component of the vector $\bar{x}$, $\hat{p}(y)$ and $\hat{p}(x^j = 1|y)$ are probabilities estimated using the training data. Several variations of Naïve Bayesian classifier have been applied for the spam filtering [8].

### B. Support Vector Machines (SVM) Classifier

Support Vector Machine (SVM) [9] has become very popular in the machine learning classifier because of its good generalization performance and its ability to handle high dimensional data by using its kernels. SVM are very effective in a wide range of bioinformatics problems.

**Figure 1: SVM Classifier**

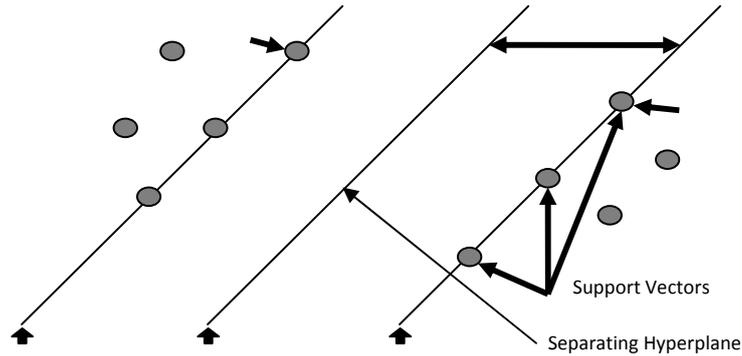

Support Vectors
Separating Hyperplane

SVM learn by example. Given the training samples and a predefined transformation $\Phi: \mathbb{R}^d \longrightarrow F,$ which maps the features to a transformed feature space, the classifier separates the samples of the two classes with a hyperplane (as shown in Figure 1) in the transformed feature space, building a decision rule of the following form:

$$f(\bar{x}) = sign\left( \sum_{i=1}^{n} \alpha_i\, y_i K(\bar{x}_i, \bar{x}) + b \right),$$

were, $K(\bar{u}, \bar{v}) = \Phi(\bar{u}).\Phi(\bar{v})$ is the kernel function and $\alpha_i, i = 1 \dots n$ and $b$ maximize the margin of the separating hyperplane. The value -1 corresponds to $C_{leg}$ and 1 corresponds to $C_{spam}$. SVM was proposed in particular for classification of vectors of features extracted from images [10]. It has undergone an improvement called

Relaxed Online SVM [11] in which the computation cost of updating the hypothesis is greatly reduced by training only on actual errors. Another improvement of this method is suggested in [12] which improve the accuracy by using locality in the spam phenomenon.

*C. Term Frequency-Inverse Document Frequency (TF-IDF) Classifier*

The name Term Frequency-Inverse Document Frequency (TF-IDF) applies to a term weighting scheme. The term weights are real numbers indicating the significance of terms in identifying a document. Based on this concept, the weight of a term in an e-mail message can be computed by the $.idf$ . The $tf$ (term frequency) indicates the number of times that a term $t$ appears in an e-mail. The $idf$ (inverse document frequency) is the inverse of document frequency in the set of e-mails that contain . The $tf.idf$ weighting classifier is defined as:

$$w_{ij} = tf_{ij}.log\frac{n}{df_i},$$

were $w_{ij}$ is the weight of $i^{th}$ term (token) in the $j^{th}$ document (message), $tf_{ij}$ is the number of occurrences of the $i^{th}$ term in the $j^{th}$ document , $df_i$ is the number of messages in which the $i^{th}$ term occurs, and $n$ as above is the total number of documents in the training set. A combination of TF-IDF and Rocchio algorithm [13] is also called TF-IDF which results in a more accurate classifier.

*D. K-Nearest Neighbor Classifier*

In Nearest Neighbor Classification examples are classified based on the class of their nearest neighbors. It is often useful to take more than one neighbor into account thus the technique is more commonly referred to as k-Nearest Neighbor (k-NN) [14] Classification where k nearest neighbors are used in determining the class. Since the training examples are needed at run-time, i.e. they need to be in memory at run-time; it is sometimes also called Memory-Based Classification. Because induction is delayed to run time, it is considered a Lazy Learning technique. Because classification is based directly on the training examples it is also called Example-Based Classification or Case-Based Classification. K−NN classification has two stages; the first is the determination of the nearest neighbors and the second is the determination of the class using those neighbors.

Let us assume that we have a training dataset $D$ made up of $(x_i)_{i\in[1,|D|]}$ training samples. The examples are described by a set of features $F$ and any numeric features have been normalized to the range [0, 1]. Each training example is labeled with a class label $y_i \in Y$. The objective is to classify an unknown example $q$. For each $x_i \in D$ we can calculate the distance between $q$ and $x_i$ as follows:

$$d(q,x_i) = \sum_{f\in F}\omega f^\delta\left(qf,x_{if}\right),$$

There are a large range of possibilities for this distance metric; a basic version for continuous and discrete attributes is:

$$d(q_{|f}, x_{if}) = \begin{cases} 0 & f \text{ discrete and } qf = x_{if} \\ 1 & f \text{ discrete and } qf \neq x_{if} \\ |q_f - x_{if}| & f \text{ continuous} \end{cases}$$

The k nearest neighbors are selected based on this distance metric. Then there are a variety of ways in which the $k$ nearest neighbors can be used to determine the class of $q$. The most straightforward approach is to assign the majority class among the nearest neighbors to the query.

It will often make sense to assign more weight to the nearer neighbors in deciding the class of the query. A fairly general technique to achieve this is distance weighted voting where the neighbors get to vote on the class of the query case with votes weighted by the inverse of their distance to the query.

$$Vote(y_i) = \sum_{c=1}^{k} \frac{1}{d(q, x_c)^n} 1(y_i, y_c),$$

Thus the vote assigned to class $y_i$ by neighbor $x_c$ is 1 divided by the distance to that neighbor, i.e. $1(y_i, y_c)$ returns 1 if the class labels match and 0 otherwise. Here $n$ would normally be 1 but values greater than 1 can be used to further reduce the influence of more distant neighbors.

### E. BART Classifier

Bayes Additive Regression Tree (BART) [15] is a learning technique to discover the unknown relationship f between a continuous output $Y$ and a $p$ dimensional vector of inputs $x = (x_1, \ldots \ldots, x_p)$. Rather than using a single regression tree, BART uses a sum of tree models that can account for additive effects. By applying Bayesian learning, BART can use newly coming data to update the current model instead of re-filling the entire model. Research work carried in [16] suggests and explains the use of BART for Spam classification.

### F. Other Classifiers

Various other learning classifiers include Fisher's linear discriminator, Logistic regression, Perception, Quadratic classifiers, Boosting, Decision trees, Random forests, Neural networks, Bayesian networks, and Hidden Markov models.

### 3. Design Challenges for Spam Filter

Filtering techniques that filter spam at the receiving client are easier to deploy and monitor; however they cannot prevent spam from misusing network bandwidth and storage resources and thus are considered to be least effective techniques. Filtering techniques that filter spam on the receiving server can stop spam even before it is stored locally. Further, they can make better decisions, by aggregating information across multiple spam recipients. Enterprise solutions can detect messages that are delivered to multiple users and that are likely spam, or can even connect to centralized repositories of spam information. Such repositories can be all computer constructed (based on individual message scores) or can include ratings by trusted human users. Filtering at the receiving end does nothing about the costs of spam delivery and storage, which are still borne by the receiver. The reduction in spam throughput itself is not sufficient to provide a strong negative feedback loop that will deter spammers from their activities. Instead, there is a positive feedback loop, as spammers send larger spam volumes to compensate for the reduced throughput due to filtering. It is because of this reaction of spammers that we receive several copies of the same spam message which indicates spammers' indiscriminate use of various e-mail lists they has been acquired by them. False positives (valid messages flagged as spam) continue to occur as long as our contacts list continues to expand, for example through face to face or phone meetings. Information about these new contacts is relevant for spam filtering, but is not automatically available to spam control tools. Similarly, changes in likes or dislikes are difficult to update into the settings of spam control tools.

In response to improvements in filtering techniques, spammers develop new techniques to fool the anti-spam procedures and reduce their effectiveness [17]. Various attacks that have been reducing the efficiency of spam filters include tokenization attack, obfuscation attack and statistical attacks. In the tokenization attack, spammers intend to prevent correct tokenization of the message by splitting or modifying features, for example putting extra spaces in the middle of the words. In the obfuscation attacks spammers obscure the filter in different ways for example by means of encoding. The statistical attacks skew the message's statistics for example by a good word attack. Besides these groups of attackers, a spammer may minimize the length and content of e-mail thereby removing or sufficiently reducing the context of the message so as to confuse the filter. Also spammers avoid use of any contentious wording or insert random letters and numbers into words and phrases. Some spamming attacks include use of personalized information obtained by directory harvesting attacks and thus creating personalized spam messages which directly address the recipient and often adopt spam format that is a normal business format. Such spam messages often include a greeting message to confuse the filter.

## 4. Evaluation Metrics for Spam Filters

The development of a new filter can be simplified by some existing software tools like Spamato system [18]. These tools provide a uniform user-friendly software framework and simplify practical implementation of new filters and filtering algorithms. Further, they provide and E-mail Mining Toolkit (EMT) [19] which is a data mining toolkit designed to analyze offline e-mail corpora. To evaluate the predictive accuracy of classifiers several measures have been proposed in literature [20]. The most simple measure is filtering accuracy namely percentage of messages classified correctly [21]. More informative measures are recall and precision. Spam recall measures the percentage of spam messages that the filter manages to block (filter's efficiency). Spam precision measures the degree to which the blocked messages are indeed spam (filter's safety). Research work carried out in [22] proposed weighted error rate and weighted accuracy as a measure to evaluate filter accuracy. TCR is the relative cost of using the filter (and so having some false positives and some false negatives) to using no filter at all (and so having all the spam misclassified, but all the legitimate mail classified correctly). F-measure is the weighted harmonic mean of precision and recall. Since false positive are often more expensive than false negative, it is vital to compare the false positive rate of the classifier. The Receiver Operator Characteristics (ROC) curve is a graph to plot false positive against true positive, in which various threshold values are compared.

Let $n_{L \to L}$ be the number of legitimate messages classified as legitimate, $n_{L \to S}$ be the number of legitimate messages misclassified as spam, $n_{S \to S}$ be the number of spam messages classified as spam, $n_{S \to L}$ be the number of spam messages misclassified as legitimate and $\lambda$ be the weight on the accuracy of the classifier. Then various filter measures can be calculated with formulas mentioned against each as specified in table1.

*Table 1: Evaluation measures for Spam Filters*

| Evaluation Measure | Evaluation Function |
|---|---|
| Accuracy | $Acc = \dfrac{n_{L \to L} + n_{S \to S}}{n_{L \to L} + n_{L \to S} + n_{S \to L} + n_{S \to S}}$ |
| Error Rate | $Err_{Rate} = \dfrac{n_{L \to S} + n_{S \to L}}{n_{L \to L} + n_{L \to S} + n_{S \to L} + n_{S \to S}}$ |
| False Positive Rate | $FP_{Rate} = \dfrac{n_{L \to S}}{n_{L \to L} + n_{L \to S}}$ |
| False Negative Rate | $FN_{Rate} = \dfrac{n_{S \to L}}{n_{S \to L} + n_{S \to S}}$ |
| Recall | $r = \dfrac{n_{S \to S}}{n_{S \to L} + n_{S \to S}}$ |
| Precision | $p = \dfrac{n_{S \to S}}{n_{L \to S} + n_{S \to S}}$ |

| Evaluation Measure | Evaluation Function |
|---|---|
| Weighted Accuracy | $W_{Acc(\lambda)} = \dfrac{\lambda.n_{L \to L} + n_{S \to S}}{\lambda.(n_{L \to L} + n_{L \to S}) + n_{S \to L} + n_{S \to S}}$ |
| Weighted Error Rate | $W_{Err(\lambda)} = \dfrac{\lambda.n_{L \to S} + n_{S \to L}}{\lambda.(n_{L \to L} + n_{L \to S}) + n_{S \to L} + n_{S \to S}}$ |
| Total Cost Ratio | $Tc_{Ratio} = \dfrac{n_{S \to L} + n_{S \to S}}{\lambda.n_{L \to S} + n_{S \to L}}$ |
| F-measure | $f1 = \dfrac{2pr}{p + r}$<br>(With $r$ and $p$ evenly weighted) |
| ROC Curve | True positive rate plotted against false positive rate |

## 5. Experimental Setup and Results

In order to test the performance of spam filters based on four different classifiers namely Baye's Additive Regression Tree modified for classification (CBART), Naïve Bayesian (NB), Support Vector Machine (SVM) and Neural Network (NN), we used a real life data set. This dataset was created by us from e-mail received in ten different e-mail accounts set up on different mail servers for the said purpose. This dataset was chosen to test the filters in real-life conditions having an advantage of working with up to date data. The number of spam messages in the dataset was 3300 and the total number of legitimate messages was 4700 with no messages having its duplicates in the dataset. We, in our experiment used different data sizes with different number of features.

Table 2 shows the accuracy of filters with respect to different data sizes. It can be noted that the accuracy of all filters improved with the increase in data size. Accuracy of Naïve Bayesian classifier improved by over 2% while that of CBART classifier improved by approximately 4%. The accuracy of SVM and NN classifiers based filters improved by less than 2%. Further, it was observed that the accuracy of NB based filter improved greatly for higher percentage of spam.

*Table 2: Filter Accuracy on Data Size*

| Data Size | Spam % | NB | SVM | NNet | CBART |
|---|---|---|---|---|---|
| 1500 | 33% | 94.67% | 91.93% | 90.00% | 92.00% |
| 3000 | 43% | 95.13% | 90.67% | 87.50% | 92.50% |
| 4500 | 66% | 97.11% | 92.89% | 91.78% | 94.22% |
| 6000 | 45% | 96.25% | 93.17% | 90.17% | 96.10% |
| 8000 | 45% | 96.69% | 92.85% | 91.70% | 96.94% |

In table 3 the accuracy of filters with respect to different feature sizes is shown. The accuracy of all filters improved with increase in the feature size of the classifier. For SVM and NN based filters accuracy improved by10% which is higher than that of filters based on NB and CBART classifiers by 4%. Further, NB and CBART based filters provided accuracy over 90% for all feature sizes.

*Table 3: Table 2: Filter Accuracy on Number of Features*

| Number of Features | NB | SVM | NNet | CBART |
|---|---|---|---|---|
| 10 | 91.43% | 82.49% | 81.23% | 90.23% |
| 25 | 93.98% | 86.34% | 85.43% | 92.56% |
| 45 | 95.40% | 89.45% | 86.56% | 94.43% |
| 50 | 96.69% | 92.85% | 91.70% | 96.94% |

Performance of filters in terms of precision, recall, false positive rate, false negative rate, and F-measure for the filters based on four different classifiers is shown in table 4. This comparison is based on a data size of 8000 e-mails using 50 features for classification. In table 5, the weighted error rate obtained for various filters using different weight is compared.

*Table 4: Performance in terms of Precision, Recall, FP, FN and f1 of various classifiers*

| Measure | NB | SVM | NNet | CBART |
|---|---|---|---|---|
| Precision ($p$) | 96.05% | 89.79% | 87.92% | 96.22% |
| Recall ($r$) | 95.91% | 93.27% | 92.61% | 96.36% |
| False Positive Rate ($FP_{Rate}$) | 3.95% | 10.21% | 12.08% | 3.78% |
| False Negative Rate ($FN_{Rate}$) | 4.09% | 6.73% | 7.39% | 3.64% |
| F-measure ($f1$) | 90.20% | 91.50% | 90.20% | 96.29% |

Table 5: Weighted Error rate with different values for $\lambda$.

| Weight ($\lambda$) | NB | SVM | NNet | CBART |
|---|---|---|---|---|
| 1 | 3.31% | 7.15% | 8.30% | 3.06% |
| 9 | 2.86% | 7.39% | 8.82% | 2.73% |
| 999 | 2.80% | 7.40% | 8.90% | 2.70% |

The results obtained from the comparative study of performances of filters based on different classification techniques showed that none of these filters obtained 100% predictive accuracy. However, CBART and NB classifier based filters achieved highest 96% precision. The results also demonstrate that NN based filter produced highest false positive rate followed by SVM based filter. The false positive rate of NB and CBART based filters is observed to be less than 4% which is much less than others. The F-measure of CBART classifier based filter remained highest at over 96%. Further, from table 5 it is evident that the weighted error rate of CBART classifier based filter is lowest in comparison to that of filters based on other classifiers. This rate further reduced with higher values of weight.

Various results obtained from the experiments demonstrate that CBART and NB classifiers based filters showed better performance in terms of all measuring parameters. From within CBART and NB classifier based filter techniques, CBART classifier based filter showed slightly higher performance but at the cost of speed of filtering.

## 6. Conclusion

The detection of spam at a place close to the sending server is an important issue in the network security and machine learning techniques have a very important role in this topic. In this paper, we reviewed some machine learning techniques used in spam filters and presented challenges faced by these techniques. We also presented an empirical evaluation in terms of various metrics of four machine learning algorithms namely Naïve Bayes, Term Frequency-Inverse Document Frequency, K-Nearest Neighbor, Support Vector Machine, and Classified Bayes Additive Regression Tree. The evaluation was based on 8000 e-mail messages collected from different e-mail accounts located on different e-mail servers. Although all learning classifiers showed ability to learn but the CBART and NB classifiers based filters showed better performance in terms of all measuring parameters. However, none of these classification techniques showed 100% predicative accuracy. The dynamic structure of spam and the reaction of spammers towards spam filters makes spam filtering an active area for research and thus there exists a wide scope for development of new spam filters and improvements in the existing ones.